# Advancing the State of the Art in Open Domain Dialog Systems through the Alexa Prize

Chandra Khatri[1]   Behnam Hedayatnia[1]   Anu Venkatesh[1]   Jeff Nunn[1]
Yi Pan[1]            Qing Liu[1]             Han Song[1]         Anna Gottardi[1]
Sanjeev Kwatra[1]    Sanju Pancholi[1]       Ming Cheng[1]       Qinglang Chen[1]
Lauren Stubel[1]     Karthik Gopalakrishnan[1] Kate Bland[1]     Raefer Gabriel[1]
Arindam Mandal[1]    Dilek Hakkani-Tur[1]    Gene Hwang[1]       Nate Michel[1]
                     Eric King[1]            Rohit Prasad[1]

[1]Amazon Alexa Prize
{ckhatri, behnam, anuvenk, jeffnunn}@amazon.com
{yipan, qqliu, hasong, gottardi }@amazon.com
{kwatras, pansanju, chengmc, qinglan}@amazon.com
{stubells, karthgop, kateblan, raeferg}@amazon.com
{arindamm, hakkanit, ehwang, natmiche}@amazon.com
{kinr, roprasad}@amazon.com

## Abstract

Building open domain conversational systems that allow users to have engaging conversations on topics of their choice is a challenging task. Alexa Prize was launched in 2016 to tackle the problem of achieving natural, sustained, coherent and engaging open-domain dialogs. In the second iteration of the competition in 2018, university teams advanced the state of the art by using context in dialog models, leveraging knowledge graphs for language understanding, handling complex utterances, building statistical and hierarchical dialog managers, and leveraging model-driven signals from user responses. The 2018 competition also included the provision of a suite of tools and models to the competitors including the CoBot (conversational bot) toolkit, topic and dialog act detection models, conversation evaluators, and a sensitive content detection model so that the competing teams could focus on building knowledge-rich, coherent and engaging multi-turn dialog systems. This paper outlines the advances developed by the university teams as well as the Alexa Prize team to achieve the common goal of advancing the science of Conversational AI. We address several key open-ended problems such as conversational speech recognition, open domain natural language understanding, commonsense reasoning, statistical dialog management and dialog evaluation. These collaborative efforts have driven improved experiences by Alexa users to an average rating of 3.61, median duration of 2 mins 18 seconds, and average turns to 14.6, increases of 14%, 92%, 54% respectively since the launch of the 2018 competition. For conversational speech recognition, we have improved our relative Word Error Rate by 55% and our relative Entity Error Rate by 34% since the launch of the Alexa Prize. Socialbots improved in quality significantly more rapidly in 2018, in part due to the release of the CoBot toolkit, with new entrants attaining an average rating of 3.35 just 1 week into the semifinals, compared to 9 weeks in the 2017 competition.

## 1 Introduction

Conversational AI is one of the hardest problem domains in artificial intelligence, due to the subjectivity involved in interpreting human language. The problems associated with the Conversational AI domain include natural language understanding, knowledge representation, commonsense reasoning, and dialog evaluation. A complete solution to this problem is likely to have a system on the scale of human parity, which is hard to measure (Hassan et al., 2018; Xiong et al., 2016). Such systems can be described as AI complete. With advancements in Deep Learning and AI, we have obtained significant performance improvements on subproblems within AI-Complete areas such as Image Recognition and Computer Vision. These advancements have come in large part due to the objective nature of evaluating solutions to these problems. Conversational AI requires both natural language understanding and response generation, where the latter features a potentially unbounded response space, and a lack of objective success metrics, making it a highly challenging problem to model.

Voice assistants such as Alexa and Google Assistant have significantly advanced the state of science for goal-directed conversations and successfully deployed these systems in production. However, the challenge of building agents which can carry multi-turn open-domain conversations is still far from being solved. To address these challenges and further the state of Conversational AI, Amazon launched a yearly competition called Alexa Prize in 2016. The grand challenge objective is to build agents that can converse coherently and engagingly with humans for 20 minutes, and obtain a 4 out of 5 or higher rating from humans interacting with them. Apart from the Alexa Prize, there have been challenges like Dialog System Technology Challenge (DSTC) (Williams et al., 2016) (task based and closed domain) and Conversational AI Challenge (ConvAI) (Burtsev et al., 2018) (persona based, chit-chat and challenges with evaluation). Both of these challenges are text based as opposed to speech-based. Achieving natural, sustained, coherent and engaging open-domain dialogs in spoken form, which can be evaluated and measured for success, is the primary goal of the Alexa Prize. Through the Alexa Prize competition, participating universities have been able to conduct research and test hypotheses with real customer interactions by building socialbots, which are available to all Alexa users. Users interact with socialbots via the "Alexa, let's chat" experience, engage in a live conversation, and can leave ratings and feedback for teams at the end of their conversations, which the teams can then use to continuously evaluate and improve their systems.

The inaugural cohort consisted of 16 university teams that set the path for future research, did extensive experimentation and brought advancements across Natural Language Understanding (NLU), Knowledge Ingestion, Dialog Manager, Personalization, Response Generation and Selection (Ram et al., 2017). The winning team - Sounding Board (Fang et al., 2018) from the University of Washington achieved an average rating of 3.17 and an average conversation duration of 10 minutes and 22 seconds during the finals of the inaugural year of the competition.

The 2018 cohort was made up of 8 teams, with 5 of them alumni of Alexa Prize 2017. 2018 teams not only expanded upon the learnings and research done in 2017 but also built several new components to handle compound sentences such as "I love watching the Harry Potter movies, but I think the books are more enjoyable" as well as contextual models for multi-turn dialogs. They also leveraged topic models and dialog acts for driving deep topical conversations and adapted their dialog manager based on user's interest. Teams also utilized extensive knowledge bases to enable broad coverage of topics and entities, and used a variety of linking techniques to transition between topics and entities within a knowledge graph.

Based on the learnings from the 2017 competition, we provided several resources, models and tools to the university teams for 2018. To help teams focus more on science work and minimize effort spent on infrastructure for model hosting and scaling, we created CoBot, a conversational bot toolkit in Python for natural language understanding and dialog management. We also shared a variety of models such as Topic and Dialog Act Classifiers, Conversation Evaluators, and Sensitive Content Detector, which were broadly adopted by many teams. Furthermore, we drastically improved our Conversational Speech Recognition system, with a 25% relative improvement in Word Error Rate (WER) since the end of last year, to minimize errors in inputs to the socialbots. We also provided teams with weekly metrics reports computed from human annotations and statistical models to identify potential growth areas for each.

While there are a few datasets and challenges targeting improved dialogs and conversational AI, most of them are task oriented (e.g. DSTC Challenge) or targeted at more phatic or chatty interactions (e.g. ConvAI Challenge). The Alexa Prize is very different as it addresses many of the missing gaps pertaining to task-based datasets/challenges and chitchat or persona-based conversations. The conversations are spoken, not task-restricted, open-ended, topical, involve opinions, and are conducted with real users with continuous feedback. This enables the system to evolve over time, perform A/B tests and ingest learnings in real-time. Evaluation is completely subjective and the experience is rated by a human in the loop. This data is extremely useful not only for improving existing systems such as conversational speech recognition and dialog act detection, but also for a variety of open research problems such as dialog evaluation, statistical dialog management, and sensitive content detection.

In 2017, during the inaugural Alexa Prize participants performed a significant amount of research around a variety of models, primarily focused on natural language understanding, knowledge ingestion, dialog and context modeling, personalization, response generation and ranking and response selection. This work is described in more detail in the 1$^{st}$ Proceedings of the Alexa Prize.

In this paper, we describe the scientific and engineering advancements brought by the university teams and by Amazon to advance the state of Conversational AI during the 2018 competition. We also share the performance of socialbots across a variety of metrics including ratings, duration, coherence, engagement, and response quality. Finally, we address some of the open-ended problems in Conversational AI that we have made progress in this year, and share our learnings from the 2018 competition.

## 2   Alexa Prize Background

Similar to the 2017 Alexa Prize, upon receiving a request to engage in a conversation with Alexa, e.g. "Alexa, Let's Chat", Alexa customers were read a brief message, then connected to one of the 8 participating socialbots. Customers were provided instructions on how to end the conversation and provide ratings and feedback. The introductory message and instructions changed through the competition to keep the information relevant to the different phrases. After exiting the conversation with the socialbot, which the user could do at any time, the user was prompted for a verbal rating: "How do you feel about speaking with this socialbot again?", followed by an option to provide additional freeform feedback. Ratings and feedback were both shared back with the teams to help them improve their socialbots.

The Alexa Prize 2018 was launched to a cohort of Amazon employees on April 10, followed by a public launch on May 16 at which time all US Alexa customers could interact with the participating socialbots. We followed a similar pattern to 2017 and followed this up with a semifinal phase that ran from July 2, 2018 - August 15, 2018 where customer ratings were used to determine finalists for Alexa Prize 2018. This was followed by a feedback phase for finalists, leading up to a closed door finals on Nov 6-7, 2018. Throughout the competition, the teams were required to maintain anonymity in their interactions to ensure fairness in the competition.

To drive maximum feedback to the teams and improve user engagement, the Alexa Prize experience was promoted through Echo customer emails, social media, blogs, and third-party publications. During each phase of the competition we drove at least one significant promotion, with two during the semifinals - one at the beginning and the other towards the end. Each of these promotions were carefully timed to bring the Alexa Prize to the top of customer's attention, invite those unfamiliar to engage with the skill, and educate customers about the Prize. Especially as teams faced elimination at the end of semifinals, we strove to ensure teams had enough customer interactions to implement changes to their socialbots. This year, to drive further engagement we announced the three finalists via a live Twitch broadcast on August 30th. The event was promoted in advance and saw a wide variety of audience including academia, media, developers, and hobbyists alongside all the participating teams.

Over the course of the 2018 competition, we have driven over 60,000 hours of conversations spanning millions of interactions, 50% higher than we saw in the 2017 competition.

## 2.1 2018 Alexa Prize Finals

The finals judging event was held in Seattle November 6-7, 2018. Each interaction involved 1 person talking to the socialbot and 3 people judging independently, determining the conversation stopping point based on coherence and engagement. An indication by a majority of the judges to end the interaction was counted as the end time for interaction. Conversations were rated by the judges on a scale from 1 to 5 stars, post indication of a need to end the conversation by a majority of the judges. The finals involved multiple interactors, with each interactor having multiple conversations with each socialbot. During the finals, the teams demonstrated an improved ability to listen to what the user was saying, better entity disambiguation, and sustain more multi-turn interactions on a topic. They also demonstrated the need to develop improved capabilities for multi-intent parsing and management of dialog resulting from long, multi-intent inputs as well as the addition of more variability to their responses.

# 3    Engineering Advancements Through the Conversational Bot Toolkit (CoBot)

## 3.1 Background

During the 2017 Alexa Prize, each of the competing teams built their own unique conversational agents. In our examination of each team's work, we identified common successful strategies, as well as common pitfalls, for the purpose of providing competitors in the 2018 Alexa Prize a starting point from which to build more sophisticated agents.

To help teams focus more on advancing the state of science in Conversational AI and minimize effort spent on infrastructure hosting and scaling, we created CoBot, a conversational bot toolkit in Python for natural language understanding and dialog management. The toolkit provides a set of tools, libraries and base models designed to help develop, train and deploy open-domain or multi-domain conversational experiences through the Alexa Skills Kit (ASK). The primary goal of CoBot is to drive improved quality of conversational agents by providing a toolkit that is modular, extensible, and scalable, and provide abstractions for infrastructure and low-level tasks. CoBot provides a continuous integration pipeline where experimentation in language understanding, dialog management techniques, and response generation strategies can be integrated and tested by a single command. This enables seamless scaling from development and test to production workloads on AWS.

CoBot uses many of the same principles found in the Node.JS, Python, and Java SDKs of the Alexa Skills Kit or ASK (Kumar et al., 2017), as well as general dialog toolkits like Rasa (Bocklisch et al., 2017) and DeepPavlov (Burtsev et al., 2017). CoBot exposes generalized dialog management, state tracking, and Natural Language Understanding (NLU) capabilities that are modular in nature. Unlike other toolkits, CoBot places an emphasis on infrastructure to host models and handle massive scale at runtime. While sharing key elements with the above-referenced Alexa SDKs, CoBot focuses on open-domain or multi-domain conversational experiences, with an emphasis on integration of pre-trained and university-developed deep learning models.

## 3.2 Impact

As a result of CoBot use, new Alexa Prize 2018 teams were able to build basic socialbots in significantly reduced timeline (weeks in 2018 vs. months in 2017), and drastically cut engineering overhead from scaling efforts (<2 days to address load testing failures in 2018 vs. 1-2 weeks in 2017). Students and faculty of CoBot-based teams became active participants in the development process, not only driving our engineering efforts through feature requests, but also contributing code for enhancements or bug fixes.

### 3.3 Design Philosophy

We engineered CoBot to be flexible, stateful, and scalable, while providing means of fast iteration, end-to-end software testing and experimentation through continuous integration and continuous deployment. Following are several key characteristics of CoBot:

- **Flexible:** The CoBot toolkit allows the user to mix and match Natural Language Understanding(NLU), Natural Language Generation(NLG), and dialog strategies. Default CoBot implementations of these strategies are easily overridden, A/B test support is built-in. Because CoBot is non-opinionated, users are in control of NLU, selection strategies, ranking strategies, and the deep learning frameworks used to train models.

- **Stateful:** Coherent multi-turn conversations are difficult to achieve without the use of context. CoBot's state management stores the current state and state history of a conversation in AWS DynamoDB, a very fast key-value datastore. At a base level, the state is defined as the current user utterance, potential agent responses and any additional NLU attributes that were collected during a current turn in the conversation. The CoBot model of state management ensures low latency, along with consistent and reliable data access. It is flexible, allowing the user to store and retrieve any additional key-value pairs they want within the state.

- **Scalable:** During the 2017 Alexa Prize Amazon customers engaged in millions of conversations with socialbots, with large variability based on season, time of day and day of week, and occasional large spikes around marketing events. To efficiently handle this level of traffic in 2018, CoBot utilizes AWS Lambda, a serverless compute platform, and deploys large machine-learned models in Amazon Elastic Container Service (ECS). CoBot can be extended by making use of *local modules* (for light weight models) and *remote modules* (for large machine-learned models, whether team-developed or pre-trained libraries such as Spacy). CoBot can also take advantage of models hosted on AWS SageMaker, a fully-managed model training and inference service.

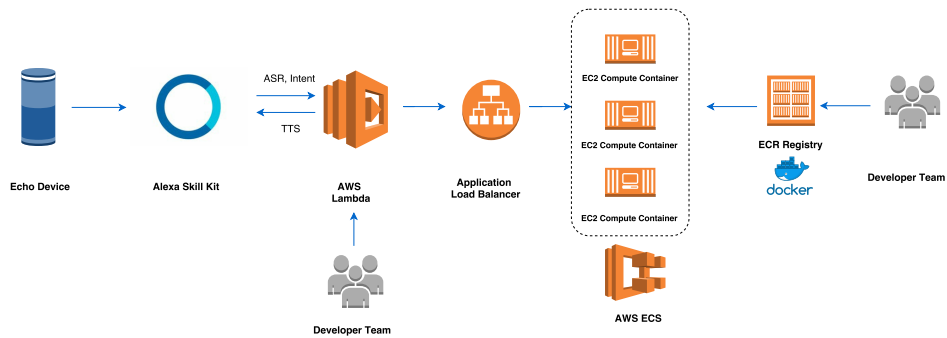

**Figure 1** CoBot System Diagram and Workflow

### 3.4 Testing and Experimentation

CoBot provides several mechanisms with which to test code or evaluate feature additions. Local end-to-end testing launches local Docker containers to test modules. The CoBot CLI can be used to simulate conversations by ingesting a text file of sample utterances, or allowing the developer to provide interactive input through the keyboard. Through the local testing, developers can iterate on code in a local development environment, before committing their changes to their Beta or Prod environments. CoBot contains robust logging, configurable with standard Python logging levels. Logs are stored in Amazon CloudWatch, a monitoring and management service, and can be used to diagnose and debug errors. Using the CoBot CLI's *transcribe* feature, a user can easily view the transcription of a conversation between the socialbot and a user.

CoBot allows experimentation by offering the ability to run A/B tests on alternately configured entry points (also called *handlers*). Users reaching one entry point may receive a different set of components, responses, or pipeline features than users reaching a separately configured entry point. The detailed architecture can be seen in Figure 2.

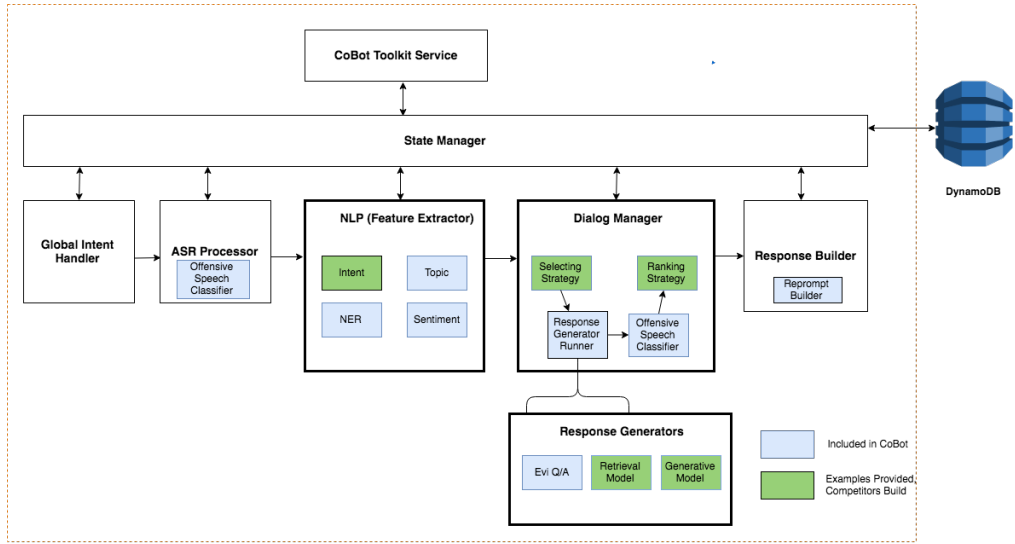

**Figure 2** CoBot Architecture

Since feedback on Alexa Prize conversations is provided partly in the form of a numeric score, CoBot developers are able to experiment with A/B testing to provide insight to which features and scientific approaches yield the best results.

**3.5 Toolkit and Data Services**

CoBot provides seamless integration to the Alexa Prize Toolkit Service, which host several deep neural network models for common conversational Natural Language Processing (NLP) tasks. This alleviates the need for developers to recreate these common models. We have included models for Topic Classification, Dialog Act Classification, Sensitive Content Detection, Question Answering and Conversation Evaluation as part of the Toolkit Service.

Because the CoBot Toolkit leverages persistent storage of data and an underlying infrastructure on AWS, we are also able to provide rich data analysis services, like data querying and visualization. Turn-based conversational and state data are persisted to DynamoDB tables. Amazon S3 buckets are used to store daily ratings and feedback data, as well as weekly coherence and engagement data, which are correlated to conversational data in DynamoDB. CoBot makes it easy to use Amazon Athena to query and analyze data directly in Amazon S3 using standard SQL commands. In addition to this, Amazon QuickSight can be used to visualize this data and create live dashboards. The ability to query and visualize real-time data makes it easy for CoBot developers to make informed and prompt decisions based on data rather than hunches.

## 4 Scientific Advancements

The Alexa Prize fosters an environment that encourages a diverse set of experimental approaches to building open domain conversational agents or socialbots. The Alexa Prize participants this year have explored novel ideas in the areas of natural language understanding, dialog management, error correction, and personalization. We highlight below some of the approaches taken by Alexa Prize teams this year in these areas. For a more in depth explanation, we refer readers to the individual team papers.

### 4.1 From The Alexa Prize Participants

#### 4.1.1 Handling Automatic Speech recognition (ASR) Errors

While performance of automatic speech recognition (ASR) has significantly improved in real-world far-field use cases such as on Alexa-enabled devices, it is still a challenging problem to achieve high accuracy across a wide array environmental conditions and for open-domain dialog systems where the user can discuss any topic and mention a huge number of unique named entities. As such, error rates tend to be higher during user dialog with Alexa Prize socialbots, and it is important for teams to be able to handle ASR errors before sending it to their socialbots for downstream tasks such as NLU processing. Iris (Ahmadvand et al., 2018) and Slugbot (Bowden et al., 2018) took a standard approach of having a threshold on word level and sentence level confidence scores. If the confidence was too low, the socialbot was prompted to ask the user to clarify what he/she said. However, if the socialbot prompts for clarification too many times this leads to a high friction experience for the user. Additionally, several teams retain n-best hypothesis for their response modules to counteract any noise in the input. Alquist took the approach of first applying case restoration techniques on ASR results before sending it to downstream applications such as NER.

Gunrock (Chen et al., 2018) took a unique approach by creating an ASR Correction module that captured homophones by leveraging a contextual knowledge base. When the word level confidence score was below a certain threshold, the socialbot would query the knowledge base for the domain of the current conversation to retrieve substitute noun phrases in the user utterance using relevant meta-phonetic (homophone) information.

#### 4.1.2 Response Modules, Ranking Strategies and Natural Language Generation

In order to deal with the vast number of possible utterances coming into a socialbot, many teams used multiple response modules where each response module would be responsible for a particular domain or set of domains. For example, one response module may handle Q/A, giving facts/news along with modules trained to respond to certain topics such as movies, music or sports. These response modules used rule-based, retrieval-based or generative techniques to produce responses. Teams used open sourced datasets or APIs for their response modules such as Reddit, News API, EVI, Wikidata, IMDB, ESPN, Washington Post, DuckDuckGo, Rotten Tomatoes, Spotify, and Bing. For rule-based bots, teams used ELIZA (Weizenbaum et al., 1966) and relied on AIML to handcraft responses. For retrieval-based methods, Alana scraped popular subreddits such as "ShowerThoughts" and "Today I Learned" and indexed them using Lucene. They then could execute a retrieval-based match for noun phrases mentioned in the user's utterance.

After generating a set of candidate responses from response modules, socialbots need to select the best response. In order to try to avoid generic and incoherent responses, Alana (Curry et al., 2018) built bot-specific classifiers using the dialogue context. Alana manually annotated responses as appropriate, inappropriate or potentially appropriate to train their model. Iris (Ahmadvand et al., 2018) selected responses for their news bot by ranking results based on how far the words in the user utterance are from the start of a news article and based on the user's preferred news domains. Eve (Fulda et al., 2018) focused more broadly on trying to identify the subset of possible responses first before narrowing down on a single response. Their method revolves around embedding conversational history into an embedding vector along with candidate responses into an embedding vector. They then looked for a candidate utterance, and shifted their conversational history embedding in that direction to match the resulting flow. SlugBot and Eve, used rankers which had hand engineered features, based on items such as confidence score from a response generator, length of an utterance, whether an utterance has already been repeated or not and if an utterance was deemed coherent.

Most teams used template based modules for rendering responses. Gunrock tried to avoid duplicate responses by creating many surface forms for each template and picking one at random. They also tried to create dynamic templates where specific slots can be swapped out. These methods, while resulting in generally coherent responses, still represent bounded simulations of the variety found in human-human conversation.

### 4.1.3 Knowledge Graphs

Representing knowledge and ingesting that knowledge in open domain conversation is a challenging problem but it is necessary in order to have multi-turn topical conversations. Many teams relied on representing knowledge in graph databases, with AWS Neptune serving as a popular choice.

Iris, Slugbot, and Gunrock used knowledge bases storing millions of different topics and entities. They leveraged open data sources such as DBPedia, Wikidata, Reddit, Twitter and IMDB. Fantom (Jonell et al., 2018) tried to take the idea further and turned their knowledge graph into a dialog structure. The graph contains nodes representing an utterance, either from the socialbot or from the user. An edge between node X to node Y means that utterance Y is an appropriate answer to utterance X. A path through the graph thus represents a dialog between user and the socialbot.

For knowledge ingestion, Alana (Curry et al., 2018) developed a system called the Contextualized Linked Concept Generator. When an entity is mentioned, they search for interesting entities within their graph using different linking strategies. Once a plausible entity in the same domain is found they will leverage the links used to find additional entities to generate a full response.

### 4.1.4 Dialogue Management

A key component of socialbots is to be able to manage dialog effectively. Many teams leveraged contextual information in their dialog management systems. This contextual information came from NLP pipelines that will be discussed in Section 4.1.7.

Many of these systems are still intent/slot based but have been extended to handle other features such as topic, sentiment, and dialog act. Alquist (Pichl et al., 2018) experimented with statistical dialogue management systems, utilizing hybrid code networks (HCN) (Williams et al., 2017) modified for the open domain use case, by taking in as input the conversational context, along with dialogue acts and system actions, and outputted either text responses or functions. Text responses are directly returned as the response while functions are represented as a reference to code which needs to be executed.

### 4.1.5 Sensitive Content Detection

Sensitive Content includes items such as offensive, abusive or hateful content. However, there are a broad set of utterances that do not have obvious profanity or hateful content but may be deemed sensitive due to recent events or deep knowledge of cultural context. To tackle sensitive content detection, teams such as Slugbot, Iris and Fantom took the standard approach by using a profanity checker before returning responses from bots. Eve (Fulda et al., 2018) tried to create sensitive detection models that utilizes context. Eve started with a blacklist of offensive words and divided it into categories of: 1) blatantly offensive words, 2) words that occur frequently in the context of offensive language 3) and words that may be offensive to an individual user which the bot only would talk about based on a user's permission.

The Alexa Prize provided a sensitive content classifier (described in Section 4.2.3) through the Cobot toolkit, which several teams utilized in their pipeline. Alana and Tartan trained their own model to detect abusive content. Tartan (Larionov et al., 2018) trained a supervised model on Reddit data using the "controversial" score given for each user message as their signal. Alana trained a model on Alexa Prize data which they annotated themselves and once abuse towards their bot was detected, they would try to drive the conversation in a different direction to mitigate further offensive content.

### 4.1.6 Customer Experience and Personalization

Iris (Ahmadvand et al., 2018) took the approach of creating a personalized topic model to predict which topic a customer would be interested in discussing next. They trained and tested a Conditional Random Field (CRF)-based sequence model on specific groups of customers: returning customers, new customers and customers split by time zones. From their experiments, they found this model has higher accuracy as compared to purely suggesting random popular topics to users. Eve tried to create a personal experience by measuring users mood through sentiment analysis. This would then be used downstream to up-vote certain response generators they believe would respond well to a certain user personality. Alana built systems to make the customer feel more welcomed. Because the conversations in Alexa Prize are highly topical there will be entities mentioned by the socialbot that a user may not have heard of. Alana implemented an entity explanation mechanism which provides domain specific information associated with a given entity, extracted from the Wikidata (Vrandečić et al., 2014) knowledge base.

### 4.1.7 Natural Language Understanding(NLU) for Open Domain Dialog

After a user has initiated a conversation, a socialbot requires a robust Natural Language Understanding (NLU) system to identify various semantic elements of a user utterance. This includes but is not limited to intent, dialog act, topic, named entities and user sentiment (Allen 1995). NLU is a difficult problem in the open domain setting because of the ambiguity and uncertainty in conversational speech. Teams described innovative approaches in areas of conversation NLU, as summarized below.

**Intent and Dialog Act Detection:** Intents represent the goal of a user for a given utterance. A dialog system needs to detect intents well to respond appropriately to that utterance. In addition to this, a conversation can be broken down and represented as a series of dialog acts that are but not limited to question, request for information, or delivery of information. Iris, Gunrock and Alquist used supervised methods to train classification models for predicting intents and dialog acts. Slugbot used the CAPC dataset provided by Alexa Prize to train their intent models.

**Named Entity Recognition (NER) and Noun Phrase (NP) extraction:** Recognizing named entities (e.g. person, organization, etc.) in an utterance is an important step to understanding a user's intent. Alquist trained custom NER models using manually labeled utterances gathered during conversations with real users. Iris and Fantom used a large database to recognize relevant entities in a user's utterance. Iris also trained an entity classifier, using DBpedia (Auer et al., 2007) entities as their dataset and a convolutional neural network.

**Anaphora and Co-reference Resolution:** Multi-turn dialog systems need to resolve ambiguous phrases with reference to prior entities for downstream tasks such as response generation. Fantom, Tartan and Slugbot did co-reference resolution by using open source tools such as SpaCy, Stanford CoreNLP, and NeuralCoref (2017). Gunrock noted that these state-of-the-art models are not adapted to conversational data and tried to build their own coreference resolution models using NER outputs. Alana performed ellipsis resolution by transforming a user utterance using contextual information, for example transforming "Yes" to "Yes I do like tea" such that their models can respond with much more detail.

**Topic Detection:** Because Alexa Prize conversations center around topics, it is crucial to predict topics of user utterances to lead to more coherent and engaging responses. Tartan, Slugbot, Fantom, Eve and Gunrock used the Cobot Topic Classifier as described in Section 4.2.2. Gunrock and Alana used knowledge graphs such as Google Knowledge graph and Microsoft Concept Graph to detect topics from named entities. Iris (Ahmadvand et al., 2018) introduced a contextual topic classifier into their system.

**Sentence Segmentation:** In natural conversation, users often express multiple thoughts in a single complex statement. Without post-processing on ASR outputs, NLU systems may face difficulty classifying these statements. Gunrock (Chen et al., 2018) trained a sentence segmentation module to break a user's utterance into to smaller segments to be able to capture the semantic meaning, and utilized start and end times of words in ASR hypotheses provided by Amazon as features to help train their model.

**Sentiment:** To better understand a user's mood, teams used sentiment detection methods. Some teams used open source tools such as VADER(Gilbert et al., 2014). Alquist trained a bidirectional recurrent (GRU) neural network on movie review data, generalizing to other conversational inputs.

## 4.2 Science Advancements from the Alexa Prize Team

### 4.2.1 Automatic Speech Recognition

Automatic Speech Recognition (ASR), a system which converts spoken words to text, is the first component a user interacts with in a spoken dialog system. It is crucial to recognize a user utterance as accurately as possible, as this will impact all of the downstream systems.

**Entity Error Rate**
Due to the highly topical nature of the Alexa Prize conversations, it is important to capture entities in user utterances which can then be used by the socialbots to return responses. In addition to measuring Word Error Rate(WER), we also actively measure Entity Error Rate(EER). Each word in a user utterance is annotated with a topic. For a particular topic say "Entertainment_Music" the number of substitution and deletion errors are counted. This metric roughly captures how well we are recognizing named entities. A key technique for reducing EER is moving to contextual ASR systems. We will present work we have done in this area taking a non-neural and neural approach for language modeling.

**Contextual Language Model Adaptation**
ASR systems traditionally consist of an acoustic model, a pronunciation model, and a statistical language model. Traditionally the role of the statistical language model is used to resolve ambiguities, regardless of context. We can improve the accuracy of the system by leveraging contextual information. For example, if language model has to determine the probability of the next word in the utterance "Alexa, I love … ". The probabilities of the next word being "The Godfather" or "The Beatles" would be very different if the user's recent interactions with Alexa were about movies or music.

We took two approaches to this problem – first, we added contextual information to a dynamic interpolation framework. In work we previously reported (Raju and Hedayatnia et al., 2018), we describe a machine-learning system that mixes different statistical language models based on contextual information. The method can leverage any type of contextual information, but in particular can examine the history a user's interactions with a socialbot system. More specifically we dynamically mix together various n-gram based language models. This is done by changing the interpolation weights for our n-gram based LM at each turn to better adapt and recognize a user utterance. We predict these interpolation weights using a Deep Neural Network that is trained to maximize the log-likelihood of the training data. We reported relative Word Error Rate (WER) reductions up to 6%. For Entity Error Rate (EER), we reduced errors by 10-15%. The EER metric is particularly important because recognizing named entities is crucial for socialbot systems. The complete list of results can be found in our results section.

We have also explored neural methods to incorporate context, involving adding contextual information for Recurrent Neural Network (RNN) models in a rescoring framework. In work presented in (Mehri et al.), we explore various methods to effectively encode conversational context into a Neural Language Model (NLM), specifically an LSTM-RNN model. Our contextual information primarily consists of the history of a user's interactions with a socialbot system. Additionally we look at integrating linguistic features such as sentence topic derived from the context. The models used to generate these derived features allow us to leverage external data sources for training. We explore various architectures on how to best incorporate this contextual information and in particular, deal with differences between socialbot responses and user utterances.

We obtained a 6.4% Word Error Rate (WER) reduction and a 7.2% Entity Error Rate (EER) reduction over a non-contextual NLM. The full results are shown in Table 4.

#### 4.2.2 Contextual Topic and Dialog Act Models

Identifying topics (such as Sports or Politics) and corresponding keywords from an utterance within a conversation helps in retrieving or generating the right set of information requested by the user. Dialog acts such as "greeting", "question", "opinion request", etc. are useful as general intents in the context of conversations, which guide the flow of the conversation. When combined, these are valuable features for open domain natural language understanding.

One of the challenges with topic, keyword and dialog act detection in open-domain conversations is effectively integrating context. In human-human or human-bot interactions, it is possible that the topic in the current utterance is a function of utterances or responses in prior turns.

To help university teams building better socialbots in 2018, we built context-aware topic and dialog act classification models. We obtained annotations for 10,000 Alexa Prize conversations with over 120,000 utterances, labeled across 12 topics (Politics, Fashion, Sports, Science and Technology, Entertainment [Music, Movies, Books, General], Phatic, Other, Interactive, Inappropriate Content) and 14 dialog acts (Information [Request, Delivery], Opinion [Request, Expression], General Chat, Clarification, Topic Switch, User Instruction, Instruction Response, Interactive, Other, Multiple Goals, Frustration Expression, Not Set). We also obtained word level annotations to obtain the topic of each word within an utterance. We trained Deep Average Network and BiLSTM based contextual models. These models are described in more detail along with some example annotations at (Khatri et al., 2018a).

#### 4.2.3 Sensitive Content Detection Classifier

In an open-domain dialog setting, one of the hardest tasks to address is detecting sensitive or offensive content. Sensitive content includes racism, profanity, hate speech, violence, sexual content or any kind of inappropriate content which may be offensive to people based on gender, demographic factors, culture or religion. There are many challenging aspects of this problem, such as coverage (a blacklist of words may not capture all kinds of sensitive content), cultural differences (sensitive content for one culture may not be sensitive for other), context (when viewed as a single utterance might seem perfectly innocuous but when viewed within a wider context they become offensive), sarcasm, non-standard vocabulary, factual correctness and recency of the content.

Most of the teams retrieve information and train models for their socialbots from publicly available data sources such as Reddit and Twitter, which are well-known to contain significant amounts of offensive materials. No large-scale labeled data set for training exists that addresses all the types of sensitive content described above.

To address this challenge, we generated training data from common internet forum conversations using a two-stage semi-supervised approach, which is illustrated in Figure 3.

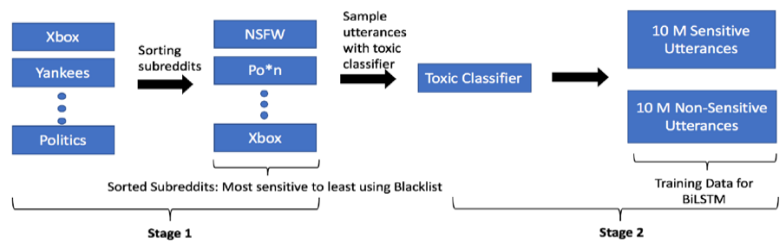

**Figure 3** Two Stage Semi-Supervision for Sensitive Content Detection

Stage 1 consists of sorting these topical forums with our blacklist by simply counting the number of words in said blacklist. Our blacklist corresponds to a manually curated list of approximately 800 offensive words. During Stage 2, high confidence sensitive comments and non-sensitive comments were sampled using a weakly supervised classifier trained on the Toxic Wikipedia Comments dataset (Jigsaw, 2018). Overall, we sampled 10 Million comments each for our sensitive and non-sensitive classes, which were then used to train a BiLSTM classifier in which word representations were initialized using GLoVE (Pennington et al., 2014) embeddings and then fine-tuned during training. More information about the model can be obtained at (Khatri et al., 2018b). Results of using this model can be found in Section 5.1.3.

### 4.2.4 Contextual Conversational Evaluators

Evaluation of dialogue systems is a challenging research problem, which despite being studied lacks a widely-agreed-upon metric. There is significant previous work on evaluating goal oriented dialogue systems such as TRAINS system (Ferguson et al., 1996), PARADISE (Walker et al., 1997), SASSI (Hone and Graham, 2000) and MIMIC (ChuCarroll, 2000), which are easier to evaluate than open-ended, non-goal oriented systems, because we can measure systems by successful completion of tasks. Venkatesh et al., 2017 describes limitations in the Turing Test (Turing 1950) model, due to divergences in information availability, objective misalignment and incentive to produce plausible but low-information content responses.

One of the primary ways in which quality of socialbots is evaluated through Alexa Prize is through user ratings. We ask Alexa users to rate the conversation on the scale of 1 to 5 based on how much they would like to speak with the socialbot again. These ratings don't provide immediate feedback during the interaction and they don't provide information about turn-level quality. To address this challenge, we defined the following five metrics for evaluating turn-level performance of open-domain dialog systems:

1. **Comprehensible:** The information provided by the socialbot made sense with respect to the user utterance.

2. **On-topic or Relevant:** The socialbot response was on the same topic as the user utterance or was relevant to the user utterance. For example, if a user asks about a baseball player on Boston Red Sox, then the socialbot should mention the correct baseball team.

3. **Response Incorrectness:** The socialbot response is irrelevant or inappropriate. In a highly subjective task, it is hard to evaluate the correctness of the response. Evaluating if the response is irrelevant or inappropriate, however, is relatively easy.

4. **Interesting:** The socialbot response contains information which is novel and relevant. For example, the socialbot would provide an answer about a baseball player and provide some additional information to create a fleshed out response.

5. **Continue Conversation:** Given the current state of the conversation and the system response, there is a natural way to continue the conversation. For example, this could be due to the socialbot asking a question to the user about the current conversation subject.

Around 15,000 Alexa Prize conversations containing 160,000 utterances were annotated by humans on the five metrics described above. We provided definitions and examples to the annotators and trained them to annotate Alexa Prize conversations. We asked them to annotate each response given the utterance and context (past turns) on how coherent and engaging the response is. They were asked to read the conversation starting from the first turn and then assign either "yes" or "no" corresponding to each metric ("is comprehensible", "is on-topic", "is incorrect", etc.) for each turn. At turn number "n", they had access to entire context in the past "n-1" turns, i.e. past utterances and responses to evaluate the response given the context.

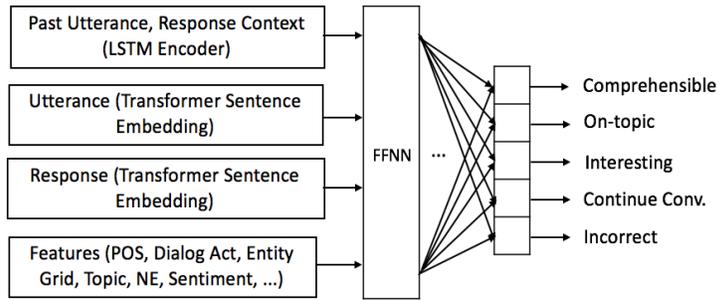

**Figure 4** Contextual Conversation Evaluators

Figure 4 depicts the model we trained corresponding to the five proposed metrics for dialog evaluation. We used several features, including context and entity-grids, to train these models. More information about the model can be obtained at (Yi et al., 2018). These models can be integrated into several dialog system components, including statistical dialog managers. More information on the metrics used in the conversation evaluators can be found in Section 5.1.4.

## 5 Results

In Section 4.2, we described several models we built and shared with the university teams for use in the Alexa Prize competition. In this section, we first provide the performance of various models that were shared with the teams and then showcase the improvement in quality of socialbots over the course of the competition on various metrics.

### 5.1 Alexa Prize Models

#### 5.1.1 Automatic Speech Recognition (ASR)

In Figure 5 we can see the relative Word Error Rate (WER) and Entity Error Rate (EER) improvement since the start of the 2018 Alexa Prize with respect to the final 2017 Alexa Prize baseline. In our currently deployed model, the WER and EER are 30% and 26% lower than at the end of the 2017 Alexa Prize (and 55% and 34% lower, respectively, than at the start of the program). Significant improvement in ASR quality have been obtained by ingesting the Alexa Prize conversation transcriptions in the models and by the Language Model advancements described in this paper.

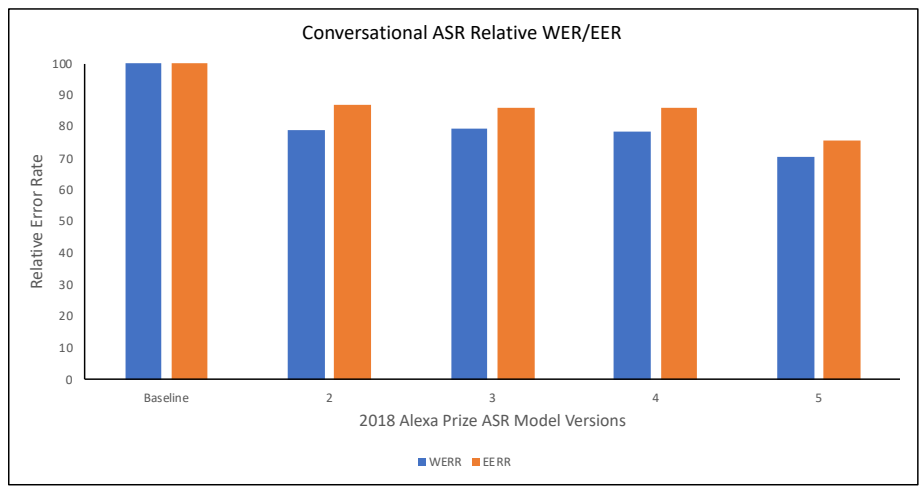

**Figure 5** Relative Word Error Rate and Entity Error Rate relative to the end of Alexa Prize 2017

**Contextual Language Model Adaptation**

We utilized two methods for contextually biasing language models – first was an n-gram based method, the results of which are described in Table 3. As described in Section 4.2.1, we bias the language model using contextual information in the form of past user utterances and TTS responses, adapting the interpolation weights of an n-gram language model at each user turn. These interpolation weights are outputted by a deep neural network (DNN) which takes in as input the contextual information. Cross-Entropy (XENT) and perplexity (PPL) refer to alternate loss functions that were used to train the DNN model. We refer the reader to (Raju and Hedayatnia et al., 2018) for more details on these models.

| Model | Features | Perplexity | Relative Word Error Rate (%) | Relative Entity Error Rate (%) |
|---|---|---|---|---|
| *Decoder: 1-pass* | | | | |
| No Adaptation (Baseline) | - | 60.77 | - | |
| DNN (Xent) | prev, meta | 59.81 | -1.73% | -8.19% |
| DNN (PPL) | prev, meta | 58.14 | -1.61% | -2.98% |
| DNN (PPL) | prev-d, meta | 55.66 | -2.76% | -10.92% |
| *Decoder: 2-pass* | | | | |
| DNN (PPL) | prev, curr, meta | 42.03 | -5.58% | -15.15% |
| DNN (PPL) | prev-d, cur, meta | 42.83 | -5.92% | -14.67% |
| DNN (PPL) | cur, meta | 42.72 | -5.98% | -15.32% |
| Topic Model | cur | 45.08 | -5.52% | -13.14% |

Table 3 Contextual ASR results

We also experimented with a neural approach to bias RNN language models, as described in Table 4. More detail on this method can be found in (Mehri et al.). "Avg Embed" and "Diff LSTM Enc" refer to methods to encode TTS responses into our model. For the first method, we average word embeddings for TTS responses and concatenate them at every time-step of our RNN model. For the second, we use a separate RNN encoder to encode our TTS responses into a hidden state vector which is used to initialize the hidden state of our decoder model.

| Models | Relative Word Error Rate (WER) | Relative Entity Error Rate (EER) |
|---|---|---|
| Context Free Baseline | - | - |
| Average Embed + Prepend | -4.70% | -5.90% |
| Different LSTM Encoder + Derived | -6.10% | -7.20% |
| Average Embed + Derived | -6.40% | -7.20% |

Table 4 Relative WER and EER using RNN Language Models w.r.t Context Free Baseline

### 5.1.2 Topic and Dialog Act Models

Based on our experiments, a bidirectional LSTM-based (BiLSTM) contextual dialog act model performs significantly better than the equivalent non-contextual model. The baseline model obtains an accuracy of 50% while adding parts of speech, context and topical features (obtained using DAN based semi-supervised topic model (Guo et al., 2017) leads to an improved accuracy of 71%. Similarly, a BiLSTM contextual topic model outperforms the non-contextual BiLSTM, Deep Average Network(DAN) and Attention Deep Average Network(ADAN) models. We obtained an accuracy of 76% for our best setting when compared to 55% obtained through our baseline model.

We have observed that these tasks are difficult for manual annotation because of the subjective nature of the problem – Table 5 provides more detailed information on inter-annotator agreement. The accuracy of the contextual dialog act (70%) and topic models (76%) is higher than the percentage of times annotation matches by all three annotators. In fact, the contextual dialog act model's accuracy is also higher than the case where at least two annotators provide the same annotation, which implies that the model has learned to generalize well on this subjective task and correct for variation in individual annotators.

For the keyword detection task, we observed that a supervised model outperforms unsupervised attention-based models (Guo et al., 2017). The contextual BiLSTM model performs better than ADAN on precision and recall of keyword detection as described in Table 6. In future work, we will explore the performance of models with explicit anaphora and co-reference resolution features. Most Alexa Prize teams used the topic, keyword and dialog act models to improve the quality of their socialbots.

| Agreement Level | Dialog Act Detection | Topic Detection |
| --- | --- | --- |
| 2 out of 3 annotators match | 70% | 89% |
| All 3 annotators match | 63% | 75% |
| Kappa Score | 41% (Moderate) | 67% (Good) |

Table 5 Inter-annotator agreement

| Supervised Keyword Detection | | | |
| --- | --- | --- | --- |
| Model | Accuracy (Binary, Multi-Class) | Keyword Precision | Keyword Recall |
| BiLSTM | 0.93, 0.57 | 0.81 | 0.55 |

Binary: Word is correctly identified as a keyword
Multi-class: Keyword class is correctly identified

| Unsupervised Keyword Detection | | |
| --- | --- | --- |
| Model | Keyword Precision | Keyword Recall |
| Attentional DAN (ADAN) | 0.37 | 0.36 |
| Contextual ADAN | 0.33 | 0.32 |
| Dialog Act + Contextual ADAN | 0.40 | 0.40 |

Table 6 Keyword detection metrics.

### 5.1.3 Sensitive Content Detection Classifier

The models trained using two-stage semi-supervision (TS Bootstrap) as described in Section 4.2.3 outperform supervised models trained on the Alexa Prize annotated test-set and on the following publicly available dataset: Toxic Wikipedia Comment data (TX). On the Alexa Prize test set (sample size: 6000 utterances), the TS Bootstrap model scored an Accuracy and F1-score of 96% and 95.5% respectively as compared to supervised models trained on annotated Toxic data, which scored 85% and 75% respectively.

We note that the model trained using large scale two-stage semi-supervision generalizes well across a variety of unseen datasets such as Twitter. Results of our Sensitive model are shown in Table 7. It can be observed that our large scale semi-supervised techniques (Bootstrap and TS Bootstrap) are able to detect challenging edge cases such as "Santa claus is the most elaborate lie ever told" and "Jerusalem is the capital of Israel".

We also note that blacklist-based approaches don't scale well - adding more blacklist words (such as "ugly" and "devil") leads to false positives while not adding them leads to poor recall. More information about the model can be obtained at (Khatri et al., 2018b).

| Model | F1-score | Accuracy |
|---|---|---|
| Blacklist *(blklist)* | 63.0 | 80.6 |
| Supervised Twitter Classifier *(sTwC)* | 74.9 | 73.9 |
| Supervised Toxic Classifier *(sToC)* | 74.0 | 85.0 |
| Blacklist based Bootstrap Classifier *(BbC)* | 92.0 | 90.0 |
| Two Stage Semi-Supervised Classifier *(tsSC)* | 95.5 | 96.0 |

| Sensitive Utterances | blklist | sTwC | sToC | BbC | tsSC |
|---|---|---|---|---|---|
| "i **** somethin' out this mornin prettier than you" | ✓ | ✓ | ✓ | ✓ | ✓ |
| "They are fat and ugly" | ✗ | ✓ | ✓ | ✓ | ✓ |
| "Santa claus is the most elaborate lie ever told" | ✗ | ✗ | ✗ | ✗ | ✓ |
| "mate, are you jewish? ******* is in the air" | ✗ | ✗ | ✗ | ✗ | ✓ |
| "why don't you get a life you sicko" | ✗ | ✗ | ✗ | ✗ | ✓ |

Table 7 Model performance and Sensitive content detection examples

### 5.1.4 Conversation Evaluator Results

Table 8 shows the results of our evaluator as compared with a baseline, which is a Deep Average Network (Iyyer et al. 2015) with utterance and response word embeddings concatenated at the input. Embeddings are fine-tuned during the task for all the models. We present precision, recall, and F-score measures along with the accuracy. Furthermore, since the class distribution of the data set is highly imbalanced, we also calculate Matthews correlation coefficient (MCC), which takes into account true and false positives and negatives. This is generally regarded as a balanced measure, useful when class sizes diverge. Evaluator predicted scores have significant positive correlation (pearson correlation of 0.2, 0.25, 0.4, and 0.3 with comprehensible, on-topic, interesting and continue conversation metrics, all with p-value of <0.0001) with the overall human evaluation score on this task.

| Evaluator | 'Yes' Class Distribution | Accuracy | Precision | Recall | F1-score | MCC |
|---|---|---|---|---|---|---|
| Comprehensible | 0.8 | 0.84 (+3%) | 0.83 (+1%) | 0.84 (+3%) | 0.84 (+8%) | 0.37 (+107%) |
| On-topic | 0.45 | 0.64 (+9%) | 0.65 (+10%) | 0.64 (+9%) | 0.64 (+12%) | 0.29 (+81%) |
| Interesting | 0.16 | 0.83 (-1 %) | 0.77 (+10%) | 0.83 (-1%) | 0.78 (+2%) | 0.12 (+inf%) |
| Conti. Conversation | 0.71 | 0.75 (+4%) | 0.73 (+5%) | 0.75 (+4%) | 0.72 (+17%) | 0.32 (+179%) |
| Incorrect | 0.44 | 0.93 (+12%) | 0.93 (+12%) | 0.93 (+12%) | 0.93 (+12%) | 0.83 (+35%) |

Numbers in parenthesis reflect the relative change when using our best model with respect to the baseline

Table 8 Conversation evaluators performance

### 5.2 Socialbot Quality

Over the course of the competition, socialbots showed a significant improvement in customer experience. In this section we provide various metrics to evaluate the quality of socialbots and evaluate the improvement in 2018 socialbot quality in comparison with that observed in the 2017 competition.

### 5.2.1 Ratings

After each conversation, Alexa users were asked to rate their interaction on a scale of 1-5. Socialbot ratings have significantly improved when compared to last year. From Figure 6, we see that from the beginning of 2018 competition, the average rating of all the socialbots in 2018 was higher than the average rating of all the socialbots in 2017 at the end of the semifinals phase. We see that the 2018 socialbots began the semifinals phase with higher ratings when compared to 2017 finalists or all 2017 socialbots. This better baseline can be explained by the fact that 2018 teams had access to the Conversational Bot Toolkit (CoBot), which had several in-built components pertaining to NLU, dialog management, and response generation as well as baseline models to start with. Furthermore, 2018 teams also had access to learnings from the architecture and techniques reported in the 2017 Proceedings of the Alexa Prize.

While the difference in average ratings across all the socialbots (until the end of semifinals) for 2018 and 2017 is significant (10%), this difference is not equally significant for the finalists (5%). There are several possible confounding factors in this year-over-year measurement – as socialbots have evolved, so did the baseline and expectations of users. To compensate for this, we surfaced a small amount of traffic to the winning socialbot in the 2017 competition (with refreshed data for 2018), and compared the ratings during the 2018 finals period. All of the 2018 finalists are rated higher than last year's winner, with 2018's top-rated socialbot more than 5% higher than last year's winner.

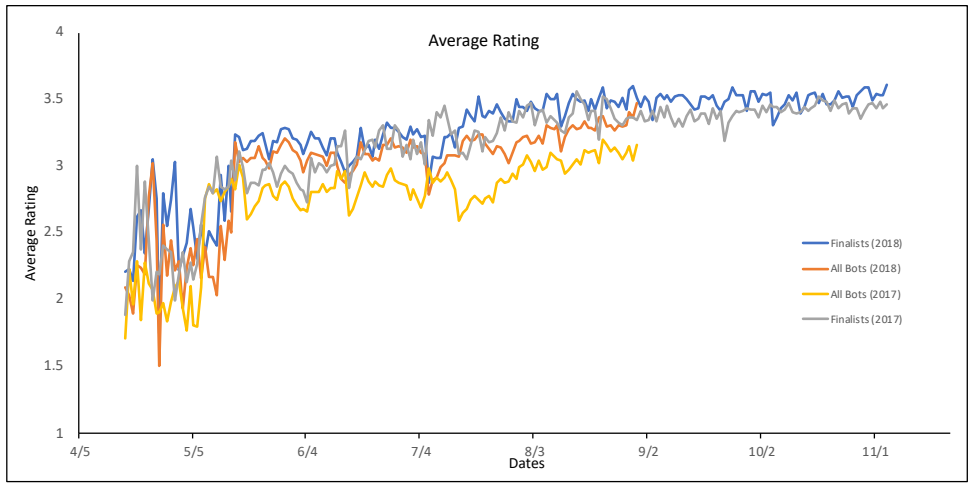

**Figure 6** Ratings provided by Alexa users to socialbots

### 5.2.2 Conversation Duration

2018 socialbots have performed significantly better on a duration basis than 2017 socialbots, as can be seen in Figure 7. By the end of the semifinals, the median duration of 2018 socialbots was 128 seconds, which is 53% higher than the 2017 socialbots at the same point in time. In fact, the median duration of all the 2018 socialbots during the end of the semifinals was 23% higher than the 2017 finalists. Overall, median duration of 2018 finalists has been around 40% higher than then 2017 finalists. We observe similar patterns with respect to the 90th percentile of the conversation duration distribution as depicted in Figure 8. 90th percentile duration for 2018 socialbots is 37% higher than that of 2017 bots for all the socialbots and for that of finalists. We consider the $90^{th}$ percentile duration as a proxy signal for the maximum plausible conversation duration of socialbot interacting with a highly interested and engaged user.

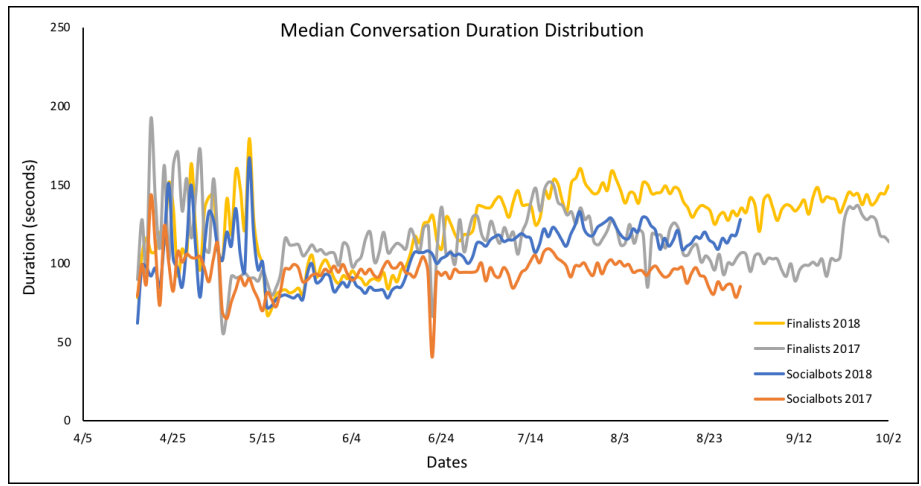

**Figure 7** Median Alexa Prize conversation duration

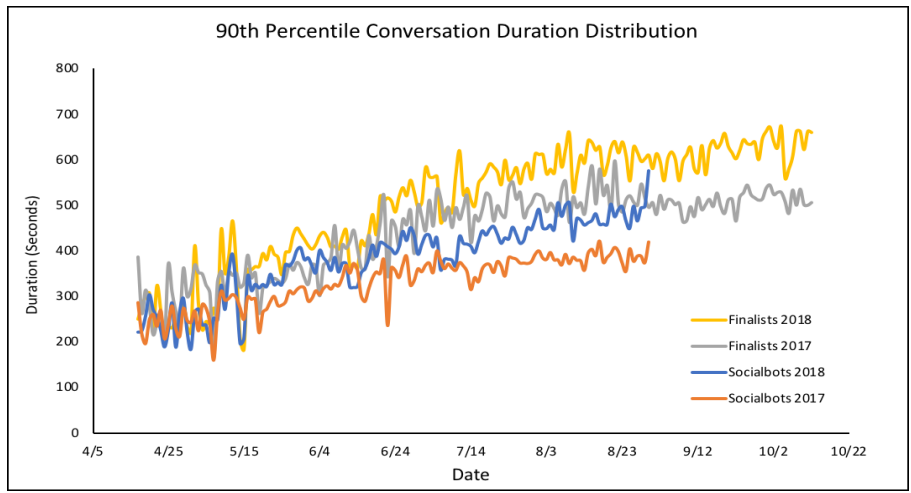

**Figure 8** Conversation Duration 90th percentile for socialbots during the competition

### 5.2.3 Number of Turns

Conversations held during the 2018 competition are significantly longer than the ones held in the 2017 competition as observed in Figure 9. The average number of turns by the end of the 2018 semifinals across all the socialbots is 12.6, which is 25% higher when compared to 2017 (10.1 turns). Two spikes can be observed in the 2017 graphs: (1) The end of semifinals and (2) The last week of September. These spikes were associated with 2017 teams introducing games and quizzes to obtain higher ratings and longer conversation – teams were instructed on both occasions to remove these, and these techniques were not permitted in the 2018 competition. Overall, the increase in the number of turns is on the order of 20-25% while the increase in conversation duration is 35-40%, which implies that 2018 socialbots are not only having longer conversation but more meaningful topical conversations with longer turn-by-turn interactions.

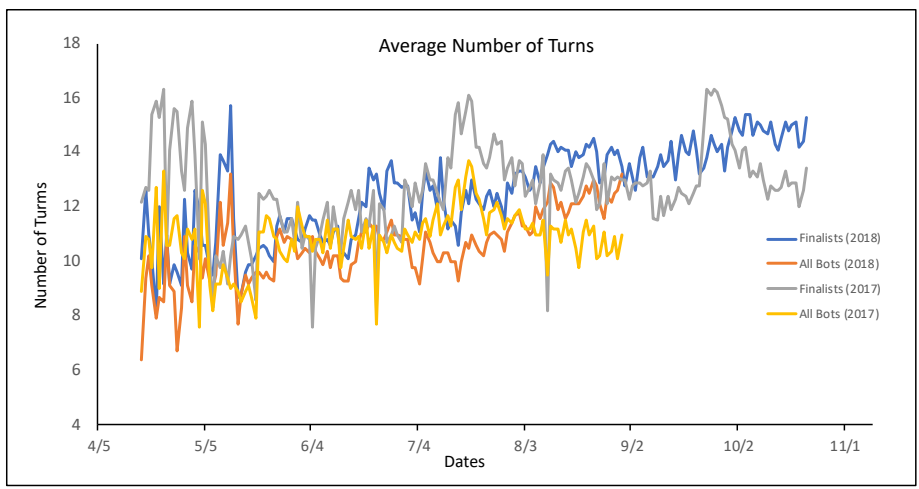

**Figure 9** Average turn distribution for socialbots during the competition

#### 5.2.4 Response Quality

We obtained annotations for response quality from the Alexa Prize conversations across all the 2018 socialbots. We defined five classes to describe the quality of a response: (1) Poor (response is not comprehensible or bot didn't understand user or response is sensitive) (2) Not Good (the socialbot had some understanding of what user said but contains incorrect or inappropriate information) (3) Passable (response is on topic but is generic or contains too little or too much of information) (4) Good (response is logical, on topic, contains accurate information but lacks personality or is not presented well and obviously coming from a bot), and (5) Excellent (contains accurate information, is complete and it is hard to identify if the response was coming from a bot). We have aggregated these classes across 3 phases: (1) Pre-semifinals (2) Semifinals and (3) Post Semifinals, as shown in Figure 10, showing that socialbots have improved through the competition cycle.

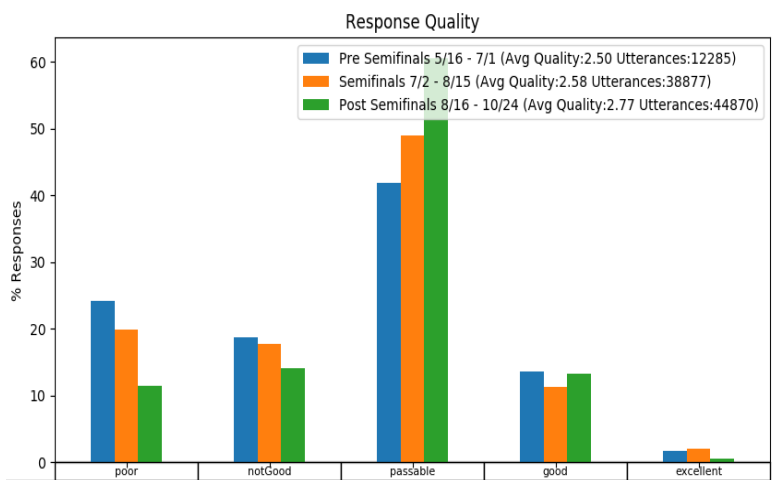

**Figure 10** Annotated Response quality for 2018 socialbots

### 5.2.5 Response Error Rate

We define Response Error Rate (RER) as the ratio of incorrect responses to total responses, as obtained through human annotations (see section 4.2 for more detail). RER is an important metric for evaluating socialbots. Given that socialbots are non-goal oriented open-domain systems, it is often hard to define a correct response - however it is easier to define and measure what is an incorrect, inappropriate or incoherent response. From Figure 11, it can be observed that the teams significantly reduced average RER through the 2018 competition. Socialbots in 2018 started with around 35% RER, but improved significantly during the semifinals stage. By the end of semifinals, the average RER for all 2018 socialbots was nearly 20% lower than the 13.22% seen at a comparable time in 2017.

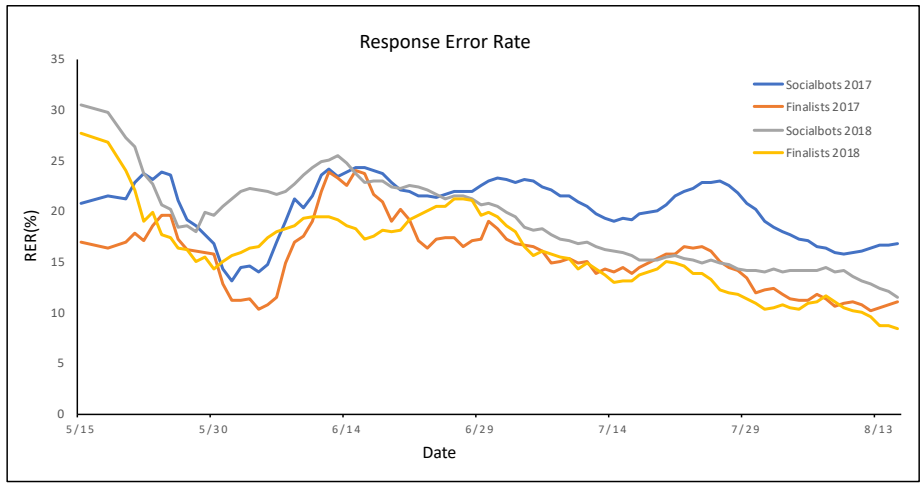

**Figure 11** Annotated Response Error Rate for Socialbots during the competition

### 5.2.6 Coherence and Engagement

Similar to response quality, we performed turn-level annotations for coherence and engagement as depicted in Figure 12. Coherence is been measured using two metrics: 1) response is comprehensible and 2) response is on topic. Engagement is measured using the following metrics: 1) response is interesting, and 2) the user would want to continue the conversation. These metrics were defined in Section 4.2 under conversation evaluators.

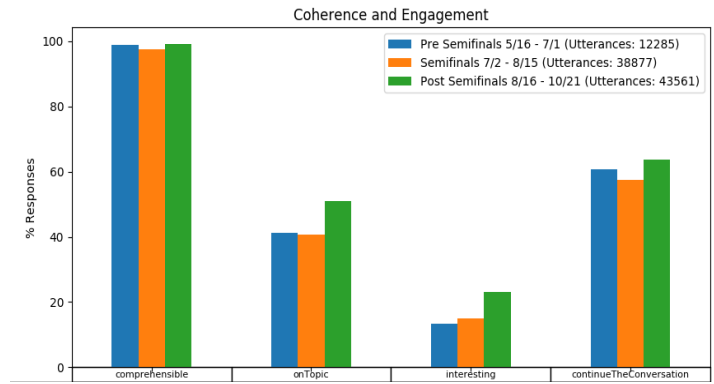

**Figure 12** Annotated Response evaluation for coherence and engagement for 2018 socialbots

We observe that socialbots have improved significantly over the competition over all the metrics mentioned above and appear to perform better in the post semi-finals stage on these metrics. We have also aggregated user ratings across all the metrics and observe that responses which are on topic and are interesting are associated with the highest ratings.

### 5.2.7 Topical Analysis

To improve the quality of socialbots, we shared a variety of data with the teams including topical analysis of conversations. From Figure 13, we can observe that "Movies", "Music" and "Technology" are the most popular topics apart from "Other". "Other" corresponds to any topic beyond the provided list of topics, as well as utterances containing multiple topics. We have observed that ratings have consistently improved across the three phases and that socialbots tend to have higher topical coverage in the post-semifinals stage compared to pre-semifinals stage, and that responses which are on the same topic lead to higher ratings.

Figure 14 illustrates that socialbots frequently tend to generate a response on the same topic as a user utterance, followed by a response classified as "Other", which is often phatic or an effort at topical redirection. This sort of topical switching behavior is associated with lower ratings, and we have previously reported a strong association between higher ratings and socialbots that can maintain longer on-topic conversations.

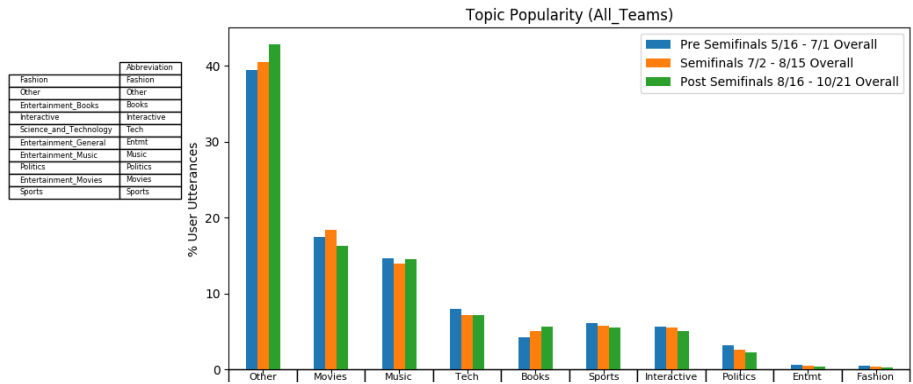

**Figure 13** Topic distribution for 2018 socialbots during three phases of the competition

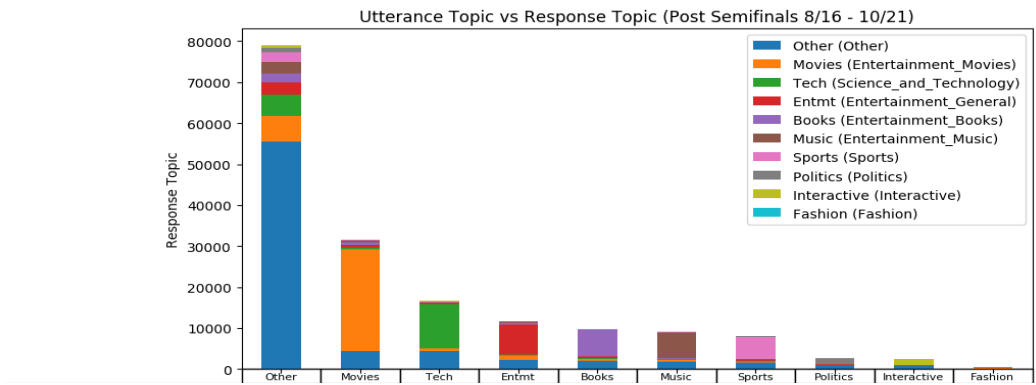

**Figure 14** Utterance vs. response topic distribution for 2018 socialbots in post semi-finals stage

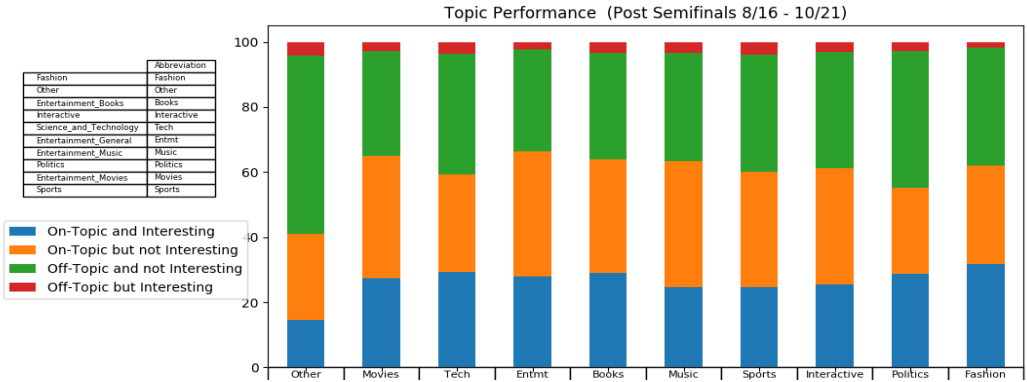

**Figure 15** Topical coherence and engagement performance for 2018 socialbots in post semi-finals stage

## 6 Discussions and Recommendations

Building an open domain conversational agent is a difficult task and there are many problems still to be solved to consistently generate coherent and engaging conversations across a wide variety of topics. From the results in Section 5.2.7 on Topical Analysis it was observed that while many responses were on topic, users did not necessarily consider them interesting. Many of these systems rely on retrieval and rule-based methods because it is hard to ingest knowledge from a structured database and present it naturally; to improve on this result, better approaches in generating relevant and engaging responses will need to be explored. Alana investigated this area with an ablation study on response modules and measured their average rating. They found that removing their rule-based module ELIZA had the largest impact on ratings. Rule/retrieval based models still play a significant role in these systems because they provide coherent responses which lead to higher ratings; however, they are relatively brittle, do not scale well to the variety of user inputs seen over a many-turn interaction, and often do not provide interesting responses for repeat users.

Additionally, the results in Section 5.2.7 on Topical Analysis showcases that socialbots are very strong in certain topics such as movies and will tend to drive the conversation towards these competency areas. We observed that socialbots tend to generate on topic responses in these strength-areas, such as Movies, but less so for areas with less robust coverage, such as Fashion. A generalized mechanism for scaling conversational response models to additional topic domains is evidently needed. Furthermore, the conversational strategy of quickly transitioning to alternate and often minimally or unrelated topical areas has been adopted by many teams to the detriment of customer satisfaction – socialbots need to be able to engage with users on a certain topic for multiple turns to sustain deep topical conversations.

Personalization is also an important factor to enhance user experience. Personalization can come in various forms such as topic selection and steering in conversation. Instead of driving conversations to areas where a bot can perform well, if conversation is driven towards topics in which users are more interested, engagement and ratings will improve. Iris did a small-scale study on how a personalized topic suggestion feature would affect ratings. From their study they found that personalized topic suggestion received an average of 4.02 rating from 360 returning users and 3.22 rating from 2,161 new users. Without personalized topic suggestion, their socialbot scored an average of 3.65 from 178 new users, while receiving only a 2.80 average rating from 52 returning users.

Finally, sentence segmentation is important to handle complex utterances. Utterances can become more complex in a conversational setting, with many pauses and hesitations in a user utterance. Unless properly parsed, it may be hard for NLU systems to understand a user's intent properly. It is also important to condition these NLU systems on contextual information such as past utterances and bot responses.

# 7 Conclusion and Future Work

Conversational AI is one of the most challenging problems in the artificial intelligence field. Several tasks associated with Conversational AI such as language understanding, knowledge representation, commonsense reasoning and dialog evaluation are believed to be "AI Complete", or truly human-intelligence equivalent problems. To address these challenges, Amazon launched the Alexa Prize competition, wherein some of the best research groups across the world work towards a common goal of advancing the state of science in Conversational AI. Alexa Prize 2018 is the second iteration of the competition and the participating university teams have successfully built socialbots which can converse with Alexa users coherently and engagingly on variety of topics. Teams built on the success of the first year of the competition with sophisticated statistical dialog management, improved personalization, complex utterance handling, greatly improved infrastructure, and the use of many machine learned models to assist in conversational language understanding.

We have observed significant scientific advancements brought by the 2018 Alexa Prize. We have also seen significant improvements in ratings, socialbot quality and conversational duration when compared to the 2017 competition. We believe we have still only touched the surface of the problem of natural human conversation, and it is still Day One for Conversational AI. We expect to see even more advancements in coming years, enabling more engaging, deeper multi-turn interactions, improved use of natural language generation in response generation, and models which can leverage both structured and unstructured knowledge sources for more robust multi-turn open-domain conversation.

**Acknowledgments**

We would like to thank all the Alexa Prize participants - university students and their advisors, for continuing to raise the bar on delivering an engaging experience to Alexa customers while pushing the boundaries of Conversational AI. We gratefully acknowledge the work of Sanghyun Yi and Alessandra Cervone, leads from Alexa Prize 2017 teams who chose to intern with the Alexa Prize team, and Rahul Goel from Alexa AI, in supporting the building of utilities that the current and future Alexa Prize participants could leverage to improve customer experience. We would also like to thank Amazon leadership for believing in the efforts being put in by the universities to further science and continuing to support the program. We additionally would like to acknowlegde the Alexa Principal Scientist community for their vision and support through this entire program; Marketing, PR, Legal and Digital Security for continuing to help drive the right messaging and a consistently high volume of traffic to the Alexa Prize skill, ensuring sufficient real world feedback for the participating universities in a secure manner. The competition would not have been possible without the support of all Alexa organizations including Engineering, Speech, NLU, Data Services and ASK and their leadership. And finally, we would like to thank Alexa customers who continue to help improve Conversational AI through their millions of interactions with the socialbots.

Pennington, J., Socher, R., & Manning, C. (2014). Glove: Global vectors for word representation. In *Proceedings of the 2014 conference on empirical methods in natural language processing (EMNLP)* (pp. 1532-1543).

Danescu-Niculescu-Mizil, C., & Lee, L. (2011, June). Chameleons in imagined conversations: A new approach to understanding coordination of linguistic style in dialogs. In *Proceedings of the 2nd Workshop on Cognitive Modeling and Computational Linguistics* (pp. 76-87). Association for Computational Linguistics.

Sutskever, I., Martens, J., & Hinton, G. E. (2011). Generating text with recurrent neural networks. In *Proceedings of the 28th International Conference on Machine Learning (ICML-11)* (pp. 1017-1024).

Iyyer, M., Manjunatha, V., Boyd-Graber, J., & Daumé III, H. (2015). Deep unordered composition rivals syntactic methods for text classification. In *Proceedings of the 53rd Annual Meeting of the Association for Computational Linguistics and the 7th International Joint Conference on Natural Language Processing (Volume 1: Long Papers)* (Vol. 1, pp. 1681-1691).

Klein, G., Kim, Y., Deng, Y., Senellart, J., & Rush, A. M. (2017). Opennmt: Open-source toolkit for neural machine translation. *arXiv preprint arXiv:1701.02810*.

Sutskever, I., Vinyals, O., & Le, Q. V. (2014). Sequence to sequence learning with neural networks. In *Advances in neural information processing systems* (pp. 3104-3112).

Curry, A. C., Papaioannou, I., Suglia, A., Agarwal, S., Shalyminov, I., Xu, X., Dušek, O., Eshghi, A.Yu, Y., & Lemon, O. (2018). Alana v2: Entertaining and Informative Open-domain Social Dialogue using Ontologies and Entity Linking. *Alexa Prize Proceedings, 2018*.

Pichl, J., Marek, P., Konrád, J., Matulík, M., & Šedivý, J. (2018). Alquist 2.0: Alexa Prize Socialbot Based on Sub-Dialogue Model. *Alexa Prize Proceedings, 2018*.

Fulda, N., Etchart, T., Myers, W., Ricks, D., Brown, Z., Szendre, J., Murdoch, B., Carr, A & Wingate, D. (2018). BYU-EVE: Mixed Initiative Dialog via Structured Knowledge Graph Traversal and Conversational Scaffolding. *Alexa Prize Proceedings, 2018*.

Jonell, P., Bystedt, M., Dŏgan, F. I., Fallgren, Per., Ivarsson, J., Slukova, M., Wennberg, U., Lopes, José., Boye, Johan & Skantze, G. (2018). Fantom: A Crowdsourced Social Chatbot using an Evolving Dialog Graph. *Alexa Prize Proceedings, 2018*.

Chen, C., Yu, D., Wen, W., Yang, Y. M., Zhang, J., Zhou, M., Jesse, K., Chau, A., Bhowmick, A., Iyer, S., Sreenivasulu, G., Cheng, R., Bhandare, A & Yu, Z. (2018). Gunrock: Building A Human-Like Social Bot By Leveraging Large Scale Real User Data. *Alexa Prize Proceedings, 2018*.

Ahmadvand, A., Choi, I., Sahijwani, H., Schmidt, J., Sun, M., Volokhin, S., Wang, Z & Agichtein, E. (2018). Emory IrisBot: An Open-Domain Conversational Bot for Personalized Information Access. *Alexa Prize Proceedings, 2018*.

Bowden, K. K., Wu, J., Cui, W., Juraska, J., Harrison, V., Schwarzmann, B., Santer, N & Walker, M. (2018). SlugBot: Developing a Computational Model and Framework of a Novel Dialogue Genre. *Alexa Prize Proceedings, 2018*.

Larionov, G., Kaden, Z., Dureddy, H. V., Kalejaiye, G. B. T., Kale, M., Potharaju, S. P., Shah, A. P., & Rudnicky, A. I. (2018). Tartan: A retrieval-based socialbot powered by a dynamic finite-state machine architecture. *Alexa Prize Proceedings, 2018*.
27